\documentclass[a4paper]{article}

\usepackage{xcolor}
\usepackage{graphicx}
\usepackage{iwslt18}
\usepackage{times}
\usepackage{latexsym}
\usepackage{amsmath}
\usepackage{multirow}
\usepackage{url}
\usepackage{booktabs}
\usepackage[utf8]{inputenc}
\usepackage{microtype}

\makeatletter
\def\blfootnote{\xdef\@thefnmark{}\@footnotetext}
\makeatother

\newcommand\witiii{WIT\raisebox{.5ex}{\small3}}

\usepackage{pifont}

\title{Pronoun Translation in English--French Machine Translation:\\
An Analysis of Error Types}

\author{Christian Hardmeier\textsuperscript{*}\qquad \\
  Uppsala University \\
  Dept. of Linguistics \& Philology \\
  Uppsala, Sweden \\
  {christian.hardmeier@lingfil.uu.se} \\\And
  Liane Guillou\textsuperscript{*}\qquad \\
  University of Edinburgh \\
  School of Informatics \\
  Scotland, United Kingdom \\
  {lguillou@inf.ed.ac.uk} \\}

\name{Firstname Lastname}
\makeatletter
\def\name#1{\gdef\@name{#1\\}}
\makeatother
\name{{\em Christian Hardmeier\textsuperscript{1*}, Liane Guillou\textsuperscript{2*}}}

\address{\textsuperscript{1}Department of Linguistics and Philology, Uppsala University\\
\textsuperscript{2}School of Informatics, The University of Edinburgh\\
{\small \tt christian.hardmeier@lingfil.uu.se, lguillou@inf.ed.ac.uk}
}

\begin{document}
\maketitle
\begin{abstract}

Pronouns are a long-standing challenge in machine translation. We present a
study of the performance of a range of rule-based, statistical and neural MT
systems on pronoun translation based on an extensive manual evaluation using the
PROTEST test suite, which enables a fine-grained analysis of different pronoun
types and sheds light on the difficulties of the task. We find that the
rule-based approaches in our corpus perform poorly as a result of
oversimplification, whereas SMT and early NMT systems exhibit significant
shortcomings due to a lack of awareness of the \emph{functional} and
\emph{referential properties} of pronouns. A recent Transformer-based NMT system
with cross-sentence context shows very promising results on non-anaphoric
pronouns and intra-sentential anaphora, but there is still considerable room for
improvement in examples with cross-sentence dependencies.

\end{abstract}

\section{Introduction}
\blfootnote{\textsuperscript{*}Both authors contributed equally.}

Pronoun translation still poses serious challenges for machine translation (MT)
systems despite years of research
\cite{LeNagard2010,HardmeierThesis,GuillouThesis,LoaicigaThesis}.  This can be
ascribed to a combination of factors including an incomplete understanding of
the problem, evaluation difficulties, and the fact that low system performance is
often obscured by the presence of many trivial problem instances. While the
generally increased fluency of neural MT (NMT) creates hope that NMT may perform
better on this task than statistical MT (SMT), this has not been convincingly
shown as yet. In this paper, we investigate the difficulties that pronouns create for
MT with the help of a detailed manual analysis of a corpus of MT output covering
rule-based, SMT and NMT systems. Our study sheds light on the problems inherent
in pronoun translation and its manual evaluation, and on the relative performance
of different types of MT systems on different types of pronouns.

Our contributions include the manual annotation and assessment of the
performance of nine English--French MT systems against the PROTEST test suite, a
comparison with the manual evaluation from the DiscoMT~2015 shared task (for a
subset of the systems), and a detailed corpus study highlighting some of the
common categories of errors revealed in a meta-evaluation of the human
judgements.
The results of our study confirm that pronoun translation remains a serious
problem for rule-based, statistical and neural MT. They strengthen previous
results indicating that rendering pronouns in translation requires modelling
both functional and referential properties \cite{GuillouThesis}, and reveal
severe weaknesses in previous modelling attempts that only addressed these
problems in part. We find that neural MT does not automatically resolve the
problem of pronoun translation. While early NMT approaches fail to outperform
SMT on pronouns, a recent Transformer-based NMT system \cite{Voita2018} achieves
very promising results for non-anaphoric pronouns and intra-sentential anaphora,
but still performs relatively poorly on examples with cross-sentence
dependencies despite explicit attempts to model inter-sentential context.



\section{The PROTEST Test Suite}
\label{sec:PROTEST}

PROTEST \cite{Guillou2016} is a test suite designed to evaluate pronouns in MT.
It consists of 250 pronouns categorised following a two-level schema.  The
pronouns in PROTEST were selected from the \textit{DiscoMT2015.test} dataset
\cite{DiscoMT2015TestSet}, a collection of TED talk
transcripts\footnote{https://www.ted.com/} and their translations. The corpus
is manually annotated with pronoun properties and links between pronouns and
their antecedents \cite{ParCor2014}.  At the top level the categories capture
\textit{pronoun function}. \textit{Anaphoric} pronouns refer to an antecedent.
\textit{Event reference} pronouns may refer to a verb, verb phrase, clause, or
an entire sentence. \textit{Pleonastic} pronouns, in contrast, do not refer to
anything. \textit{Addressee reference} pronouns are used to refer to the
reader/audience:

\hangindent=\parindent
\textit{anaphoric}: I have a bicycle. \textbf{It} is red.\\
\textit{event}: He lost his job. \textbf{It} came as a total surprise.\\
\textit{pleonastic}: \textbf{It} is raining.\\
\textit{addressee reference}: \textbf{You}'re welcome.

More fine-grained categories are derived from additional annotated features:
the pronoun's surface form, singular vs.\ plural use, subject vs.\ non-subject
position, and whether the antecedent is a group noun, an anaphoric pronoun is
inter- or intra-sentential, and an addressee reference pronoun refers to
specific people (deictic) or to people in general (generic).


\section{Data Set}
\label{sec:DataSet}

The core part of our data set consists of one rule-based MT and four SMT systems 
that participated in the shared task on pronoun translation at DiscoMT~2015
\cite{Hardmeier2015b,DiscoMT2015TestSet}. We complement this data set by
adding the output of an SMT baseline and three NMT systems.


The DiscoMT~2015 pronoun translation shared task \cite{Hardmeier2015b} studied
English--French MT, paying special attention to the translation of the English
pronouns \textit{it} and \textit{they}. Of the six submissions\footnote{One system, \textsc{a3-108} (no system
description available), was excluded from the analysis because it produced
completely unintelligible output.}, four were phrase-based SMT systems, each
trained on the shared task data, comprising Europarl \cite{Koehn2005}, News
Commentary version 9 and the shuffled news corpora from WMT 2007--2013
\cite{WMT2014}, and the \witiii\ corpus of TED talks \cite{Cettolo2012}. These four systems differ in the components that are
specific to pronoun translation. \textsc{idiap} \cite{Luong2015} and
\textsc{auto-postEDIt} \cite{Guillou2015} employ a two-pass strategy to detect
and amend incorrect pronoun translations. \textsc{uu-Tiedemann}
\cite{Tiedemann2015} does not attempt to resolve pronominal anaphora
explicitly, instead it employs a cross-sentence n-gram model over determiners
and pronouns to bias the model towards selecting correct pronouns.
\textsc{uu-Hardmeier} \cite{Hardmeier2015b} includes a neural network
classifier for pronoun prediction trained with latent anaphora resolution. The
fifth system, \textsc{its2} \cite{Loaiciga2015}, is a rule-based system
with syntax-based transfer and an anaphora resolution component
influenced by Binding Theory. An SMT baseline system implemented with Moses
\cite{KoehnMoses} is included in the data set with the label \textsc{baseline}.

\label{NMT-systems}


Following the recent shift in focus from SMT to NMT, we also assess the
performance of NMT systems on pronoun translation. We extend our corpus with
three NMT systems provided to us by researchers from leading NMT groups. The
first (\textsc{limsi}) is the \textit{s-hier} system described in paper
\cite{Bawden2018}. It was trained on OpenSubtitles2016 data \cite{OpenSubs2016},
and is designed to exploit context from previous source and target sentences
when translating discourse phenomena (including pronouns).
The second (\textsc{nyu}) is based on the NMT baseline described in
\cite{NYUNMT}. It is trained on the official training data from WMT~2014
\cite{WMT2014}. It should be noted that the \witiii\ data set of TED talks,
which can be considered in-domain for the PROTEST test suite, is not included in
the training data of the \textsc{limsi} and \textsc{nyu} systems. The third NMT
system (\textsc{yandex}) is an English-French version of a system developed for
English--Russian \cite{Voita2018}. It is based on the Transformer NMT
architecture \cite{Vaswani2017} and uses context from the preceding sentence to
improve the translation of discourse phenomena. It is trained on a subset of the
Europarl, News Commentary, and TED data from the DiscoMT~2015 shared task. We
also include the \emph{reference translation} as an upper bound to system
performance.





\section{PROTEST Annotation}
\label{sec:Annotation}

The MT system translations of the test suite pronouns were manually evaluated
using the PROTEST graphical user interface (GUI) \cite{Hardmeier2016}. The annotator is presented with the original English sentence together
with up to five sentences of context and the corresponding MT system
translation. The pronoun to be annotated is highlighted in the English
sentence, and its translation (found via source-target word-alignments) is
highlighted if available. If the pronoun is anaphoric, its antecedent
head and translation are also highlighted.
 
The translation of each pronoun in the test suite by each MT system (a
\textit{translation example}) is annotated according to the PROTEST guidelines
\cite{Hardmeier2016}. The focus of the annotation is on the pronouns and their
antecedents, which are marked as either correct or incorrect. Other words in the
translation are not evaluated, except if the translation is so bad that pronoun
evaluation is impossible. Additionally, tags are used to indicate common issues
such as \textit{bad translations} or \textit{incorrect word alignments}. The
annotations were carried out by two bilingual English-French speakers, both of
whom are native speakers of French.

\section{Inter-Annotator Agreement}

A random sample of 228 translation examples, stratified by pronoun category and
annotated by both annotators, was used to calculate inter-annotator agreement
(IAA). The remainder (2,272 examples) was divided between the two
annotators.
As the NMT systems were added
to the analysis later, no NMT examples are present in the IAA
set.

The IAA scores were calculated using Cohen's Kappa \cite{Cohen1960}.
The agreement scores are 0.71 for pronouns (\textit{good agreement}) and 0.58
for antecedents (\textit{moderate agreement}). 
The annotators disagreed on the annotation of 22/228 (12.28\%) pronouns and
8/140 (5.71\%) antecedents. An examination of the confusion matrices reveals
only one notable difference: in half of the disagreements (13 cases) annotator 1
marked a pronoun translation as correct, and annotator 2 marked it as incorrect,
possibly indicating that one of the annotators was slightly stricter.
We conclude that although the level of
agreement is reasonable, the manual annotation of pronouns and antecedents is
far from easy. The 31 examples in the IAA set for which the annotators
disagree on the annotation of the pronoun, antecedent, or both, were resolved
through adjudication by one of the authors of this study.

\section{Comparison with DiscoMT~2015 Manual Evaluation}
\label{sec:ComparisonManual}


In the DiscoMT~2015 shared task, performance was evaluated using human
judgements from a gap-filling task in which the pronoun translation in the MT
output was obscured and the annotator was asked to suggest which French
pronoun(s) would be suitable given the surrounding MT context.
The set of examples was different from PROTEST and focused only on
subject instances of \textit{it} and \textit{they}.


To compare our results with the original shared task evaluation, we first
identified a PROTEST-style category for each pronoun example in the
DiscoMT~2015 evaluation set (210 pronouns) using the manual annotations in
\textit{DiscoMT2015.test}. 
We then re-computed the official accuracy metric, named Acc+OTHER in the
DiscoMT~2015 evaluation \cite{Hardmeier2015a}, for the systems contained in the
original DiscoMT data set, restricting the pronoun examples to those with
categories matching the set used in the PROTEST evaluation (leaving 195 pronouns in
total
). Table~\ref{table:CompareManualEval} shows the Acc+OTHER
scores on the reduced set of 195 pronouns and on the 205 it/they pronouns from
the PROTEST test suite (after removing all instances of \textit{you}). For reference, we
also include the original Acc+OTHER scores from DiscoMT~2015 computed over the
full evaluation set.

\begin{table}\centering\small
\begin{tabular}{lccc}
\toprule
                       & \multicolumn{2}{c}{Compatible categories} & \textit{official}\\
                       & PROTEST                                   & DiscoMT             & \textit{DiscoMT}      \\
	               & it/they                                   & 195 ex.             & 210 ex. \\
\midrule
\textsc{baseline}      & 0.590                                     & \textbf{0.631}      & \textit{\textbf{0.676}} \\
\textsc{idiap}         & 0.600                                     & 0.595               & \textit{0.657}          \\
\textsc{uu-Tiedemann}  & \textbf{0.610}                            & 0.615               & \textit{0.643}          \\
\textsc{uu-Hardmeier}  & 0.566                                     & 0.544               & \textit{0.581}          \\
\textsc{auto-postEDIt} & 0.595                                     & 0.528               & \textit{0.543}          \\
\textsc{its2}          & 0.380                                     & 0.394               & \textit{0.419}          \\
\bottomrule
\end{tabular}
\caption{Comparison between the DiscoMT~2015 and PROTEST manual evaluations}
\label{table:CompareManualEval}
\end{table}


The system rankings are very different when Acc+OTHER on the reduced set of 195
pronouns is compared to the proportion of correctly translated pronouns in the
reduced set of 205 PROTEST pronouns. 
The \textsc{uu-Tiedemann} and \textsc{idiap} systems both beat the
\textsc{baseline} according to PROTEST rankings, and \textsc{auto-postEDIt} fares
better (cf.\ Table \ref{table:ProResults}).

To gain more insight into the differences between the evaluation methods, we
studied the examples common to both evaluations. The overlap comprised 45
examples from six MT systems, for a total of 270 judgements. We find that the
PROTEST and DiscoMT assessments agree in 205 cases (75.9\%). 30 cases (11.1\%)
differ due to annotation errors. Errors occur disproportionately often in the
DiscoMT gap-filling data set. It included 16 incorrect cases and 11 examples of
incomplete annotations, where the annotators provided correct suggestions, but
failed to list the correct pronoun selected by the MT system. The PROTEST data
set had 3 obviously incorrect annotations. The remaining 35 cases (13.0\%)
reflect problems of the annotation task, such as incorrect word alignments,
disagreement about whether to accept \textit{ils} without overt antecedents or
annotation differences in disfluent translations.

The comparison of the two evaluation procedures shows that the ranking of
systems is sensitive to the choice of examples. The original DiscoMT evaluation
used a random selection of pronoun examples designed to approximate the
distribution of pronouns in naturally occurring text. PROTEST, by contrast,
contains a stratified sample covering specific types of pronoun use. Which
selection strategy is more useful in practice must be decided from case to case
based on the purpose of the evaluation. Another noteworthy point is the high
number of annotation errors found in the DiscoMT evaluation. On the one hand,
the incomplete annotations reveal a weakness of the gap-filling evaluation
procedure, which was designed to avoid priming the annotators with the output of
the MT systems, but comes at the expense of an increased risk of missing valid
alternatives. The large number of outright annotation errors, on the other hand, 
demonstrates the dangers of using annotators that are not native speakers of the
target language. Finally, a relatively large number of disagreements seems to be
inherent in the task and shows that the evaluation of pronouns in the context of
disfluent MT output is not always a well-defined problem.

\section{MT Corpus Study}
\label{sec:MTCorpusStudy}

In this section, we present an overview of system performance based on the human
judgements collected during annotation over the PROTEST test suite.
Table~\ref{table:SupplementaryResults} details the number of correctly
translated antecedent heads and those translations which could not be evaluated
due to more general problems. Table~\ref{table:ProResults} shows a summary of
the judgements on the pronoun translations.


In addition, we manually looked through the 821 examples in the data set in the
anaphoric, event and pleonastic categories that were labelled as incorrect or
rejected due to general problems by the annotators, and created another level of
synthesis of the results by categorising the errors into different types
according to the most important problem in the translation. This meta-evaluation
was performed by one of the authors of this study, a native speaker of German
with good knowledge of English and French.

The tags assigned in this meta-evaluation are always based on a reassessment of
the examples, so the counts per category do not tally exactly with the counts of
the main analysis, even for identical tags. They fall into four broad
categories: \emph{Acceptable} means that the pronoun translations were judged to
be acceptable even though they had been rejected in the initial annotation. Most
of these examples are due to incorrect automatic word alignments between the
source and the translations. In such cases, the first-pass annotators had been
instructed not to label the pronoun translations as correct so that the
word-aligned data set could be used as a source of correct examples in other
experiments. The \emph{acceptable} category also includes some examples where
the initial annotator had missed a valid different reading (e.\,g., by
interpreting a pronoun as encoding abstract instead of concrete anaphora), but
we did not question the native speakers' acceptability judgements where no such
alternative reading was available.

The second category includes examples where the main problem was a general
mistranslation or disfluency in the translation (\emph{bad translation}), with
the omission of a large part of the sentence as a special case (\emph{missing
text}). The third category comprises translations with errors in \emph{gender}
or \emph{number agreement}, or both.

In the final category, \emph{wrong pronoun type}, the type of pronoun output by
the MT system was not compatible with the structure or semantics of the target
sentence. Typical examples are the use of a French personal pronoun like
\textit{il/elle} where a demonstrative like \textit{ce} or \textit{cela} would
have been appropriate, or vice versa. This category also includes the use of
feminine pronouns as pleonastics, or pleonastic uses of \textit{il} (acceptable
in many, but not all cases) that were judged as incorrect by our native speaker
annotators.

\begin{table}\centering\small
\begin{tabular}{lcccc}
\toprule
&\rotatebox{90}{\parbox{1.4cm}{antecedent\\[-1mm]correct}}&\rotatebox{90}{\parbox{1.4cm}{bad\\[-1mm]translation}}&
\rotatebox{90}{\parbox{1.4cm}{incorrect\\[-1mm]word\\[-1mm] alignment}}&\rotatebox{90}{\parbox{1.4cm}{missing\\[-1mm]text}}\\
\midrule
Examples               & \textit{145} & \textit{250} & \textit{250} & \textit{250} \\
\midrule
Reference              & 139          & 1            & 0            & 0 \\
\midrule
\textsc{baseline}      & 123          & 7            & 9            & 0 \\
\textsc{auto-postEDIt} & 140          & 3            & 0            & 0 \\
\textsc{uu-Hardmeier}  & 134          & 2            & 9            & 0 \\
\textsc{idiap}         & 125          & 7            & 9            & 0 \\
\textsc{its2}          & 107          & 4            & 34           & 0 \\
\textsc{uu-Tiedemann}  & 127          & 8            & 6            & 0 \\
\textsc{limsi}         & 108          & 2            & 30           & 55 \\
\textsc{nyu}           & 125          & 9            & 9            & 3 \\
\textsc{yandex}        & 135            & 2            & 12            & 0 \\
\bottomrule
\end{tabular}
\caption{Antecedent translations marked as correct, and supplementary information, per system}
\label{table:SupplementaryResults}
\end{table}

\begin{table*}\centering\small
\setlength\tabcolsep{0.25em}
\begin{tabular}{lcccccccccccccc}
\toprule
& \multicolumn{8}{c}{anaphoric} & event & pleonastic & \multicolumn{3}{c}{addressee reference} & \\ 
\cmidrule{2-9} \cmidrule{12-14}
& \multicolumn{4}{c}{it} & \multicolumn{3}{c}{they} & it/they & it & it & \multicolumn{3}{c}{you} & \\
\cmidrule{2-5} \cmidrule{6-8} \cmidrule{9-9} \cmidrule{12-14}
& \multicolumn{2}{c}{intra} & \multicolumn{2}{c}{inter} & intra & inter & sing. & group & & & generic & \multicolumn{2}{c}{deictic} & \\
\cmidrule{2-3} \cmidrule{4-5} \cmidrule{13-14}
& subj. & non-subj. & subj. & non-subj. & & & & & & & & sing. & plural & \textbf{total} \\
\midrule
\textit{Examples}                                      & \textit{25} & \textit{15} & \textit{25} & \textit{5} & \textit{25} & \textit{25} & \textit{15} & \textit{10} & \textit{30} & \textit{30} & \textit{20} & \textit{15} & \textit{10} & \textit{250} \\
\midrule
Reference                                              & 25          & 15          & 21          & 3          & 23          & 19          & 12          & 9           & 28          & 30          & 20          & 15          & 10          & 230 \\
\midrule
\textsc{baseline}                                      & 15          & 5           & 16          & 0          & 12          & \textbf{15} & 7           & 4           & 20 & 27          & 19          & \textbf{15} & \textbf{10} & 165 \\
\textsc{auto-postEDIt}                                 & 18 & 10 & 12          & 0          & 18 & 14          & 9           & 6           & 13          & 22          & \textbf{20} & \textbf{15} & \textbf{10} & 167 \\
\textsc{uu-Hardmeier}                                  & 14          & 7           & 10          & 1 & 12          & 11          & 10          & 6           & 19          & 26          & 18          & \textbf{15} & \textbf{10} & 159 \\
\textsc{idiap}                                         & 13          & 6           & 15          & 1 & 16          & 9           & 8           & \textbf{10} & 19          & 26          & 18          & \textbf{15} & 9           & 165 \\
\textsc{its2}                                          & 9           & 6           & 11          & 0          & 12          & \textbf{15} & 7           & 5           & 2           & 11          & 18          & 14          & 8           & 118 \\
\textsc{uu-Tiedemann}                                  & 15          & 3           & 15          & 1 & 13          & 14          & \textbf{12} & 4           & 19          & \textbf{29} & \textbf{20} & \textbf{15} & \textbf{10} & 170 \\
\textsc{limsi}                                         & 10          & 6           & \textbf{17} & 1 & 10          & 8           & 5           & 7           & 20 & 25          & 16          & 9           & \textbf{10} & 144 \\
\textsc{nyu}                                           & 15          & 8           & 14          & 0          & 17          & 13          & 2           & 6           & 17          & 21          & 18          & 13          & \textbf{10} & 154 \\
\textsc{yandex}                                           & \textbf{23}          & \textbf{12}           & 12          & \textbf{3}          & \textbf{21}          & 11          & 11           & 7           & \textbf{28}          & 27          & \textbf{20}          & 14          & \textbf{10} & \textbf{199} \\
\midrule
\multicolumn{14}{l}{\textit{Average over MT output}}\\
\quad count                                            & 14.7        & 7.0         & 13.6        & 0.8        & 14.6        & 12.2        & 7.9         & 6.1         & 17.4        & 23.8        & 18.6        & 13.9        & 9.7         & 160.1\\
\quad percentage                                       & 58.7        & 46.7         & 54.2        & 15.6        & 58.2        & 48.9        & 52.6         & 61.1         & 58.1        & 79.3        & 92.8        & 92.6        & 96.7         & 64.0\\
\bottomrule
\end{tabular}
\caption{Pronoun translations marked as correct, per system}
\label{table:ProResults}
\end{table*}

\begin{table*}\centering\small
\setlength\tabcolsep{0.05em}
\begin{tabular}{lccccccccccc}
\toprule
& \multicolumn{8}{c}{anaphoric} & event & pleonastic & \\ 
\cmidrule{2-9}
& \multicolumn{4}{c}{it} & \multicolumn{3}{c}{they} & it/they & it & it & \\
\cmidrule{2-5} \cmidrule{6-8} \cmidrule{9-9} 
& \multicolumn{2}{c}{intra} & \multicolumn{2}{c}{inter} & intra & inter & sing. & group & & \\
\cmidrule{2-3} \cmidrule{4-5}
& subj. & non-subj. & subj. & non-subj. & & & & & & & \textbf{total} \\
\midrule
   acceptable       & 16 & 19 & 14  & 4  & 12  & 18  & 9  & 5  & 7   & 6  & 110\\\addlinespace
   bad translation  & 21 & 37 & 29  & 21 & 18  & 12  & 27 & 7  & 20  & 27 & 219\\
   missing text     & 11 & 5  & 2   & -- & 8   & 6   & 7  & 1  & 7   & 4  & 51 \\\addlinespace
  gender agreement  & 43 & 6  & 50  & 13 & 53  & 77  & 1  & 7  & 1   & 3  & 254 \\
  number agreement  & 1  & 1  & 1   & -- & 7   & 4   & 14 & 5  & --  & -- & 33 \\
  gender and number & -- & -- & 1   & -- & --  & 2   & 9  & 2  & --  & 1  & 14 \\\addlinespace
wrong pronoun type  & 5  & 8  & 16  & 2  & 2   & 2   & -- & 9  & 80  & 15 & 139 \\\addlinespace
total               & 97 & 76 & 113 & 40 & 100 & 121 & 67 & 36 & 115 & 56 & 821\\
\bottomrule
\end{tabular}
\caption{\label{tab:synth-categories}Meta-evaluation: Common error sources}
\end{table*}

\begin{table*}\centering\small
\newcommand\R[1]{\rotatebox{90}{#1}}
\newcommand\RR[1]{\rotatebox{90}{\textsc{#1}}}
\begin{tabular}{lrrrrrrrrrrr}
\toprule
&\R{Reference}&\RR{baseline}&\RR{idiap}&\RR{uu-Tiedemann}&\RR{uu-Hardmeier}&\RR{auto-postEDIt}&\RR{its2}&\RR{limsi}&\RR{nyu}&\RR{yandex}&\R{total}\\
\midrule
   acceptable       & 14 & 12 & 6  & 8  & 8  & 6  & 27  & 8  & 12 & 9  & 110\\\addlinespace
   bad translation  & 1  & 28 & 31 & 32 & 23 & 22 & 24  & 12 & 34 & 12 & 219\\
      missing text  & -- & -- & -- & -- & -- & -- & --  & 50 & 1  & -- & 51\\\addlinespace
  gender agreement  & 5  & 28 & 29 & 27 & 40 & 30 & 30  & 17 & 25 & 23 & 254\\
  number agreement  & 2  & 6  & 1  & 3  & 4  & 1  & 6   & 3  & 5  & 2  & 33\\
  gender and number & 1  & 1  & 4  & -- & 1  & 2  & 4   & -- & 1  & 1  & 15\\\addlinespace
wrong pronoun type  & 2  & 12 & 12 & 12 & 14 & 22 & 39  & 7  & 16 & 3  & 139\\\addlinespace
total               & 25 & 87 & 83 & 82 & 90 & 83 & 130 & 97 & 94 & 50 & 821\\
\bottomrule
\end{tabular}
\caption{\label{tab:synth-systems}Meta-evaluation: Error types per system}
\end{table*}

\subsection{Comparison of Functional Categories}

Table~\ref{tab:synth-categories} shows the most common error types found in the
data for each of the pronoun categories. Note that the examples in the test
suite were the same for all systems and many of the systems made similar errors
on the same examples. This should be kept in mind when interpreting the numbers
in this table.

In theory, a translation can also be grammatical while not containing a direct
translation of the source pronoun. 
Whilst we found very few examples of this type in the dataset, we should stress that the
rareness of such examples in the dataset does not imply that this is not a
relevant translation strategy. 
Rather, this type of alternation is beyond the
capabilities of current MT systems, and examples where it would have been
appropriate are likely to have ended up in another error category.

Clear patterns are visible in Table~\ref{tab:synth-categories}. In the
\textbf{anaphoric \textit{it}} categories, there is a split between subject and
non-subject pronouns. \emph{Bad translation} is very common for non-subject
pronouns, and this frequently means that the pronouns are simply omitted bby the
MT systems. This happens far less frequently for subject \textit{it}, where
\textit{gender agreement} is the most important problem, followed by
\textit{wrong pronoun type}, which in this category usually means that
\textit{ce} or another demonstrative was chosen where a personal pronoun would
have been appropriate. The patterns for inter- and intra-sentential \textit{it}
are very similar, but the inter-sentential category has more \textit{wrong
pronoun type} cases. One possible explanation is that in the inter-sentential
examples sentences often start with \textit{it's}, inviting confusion with
pleonastic \textit{it}. The context of the intra-sentential examples tends to
include clues like content-bearing verbs that disambiguate the pronoun function
more strongly.


In the \textbf{intra- and inter-sentential anaphoric \textit{they}} categories,
gender agreement is also the most prevalent source of errors, especially in the
inter-sentential case. This is not unexpected, as gender marking of plural
pronouns is known to be a challenging problem for MT to such an extent that
accuracy on these cases has even been used as a benchmark metric in previous
work \cite{Hardmeier2013a}. There are also some cases of number agreement
errors. In the intra-sentential category, most of these stem from a single
example in the test suite. It involves the pronoun \textit{someone}, which is
correctly referred to with singular \textit{they} in English, but requires
singular agreement in French. The number agreement examples in the
inter-sentential category are spread over different examples, and we could not
discern any clear patterns.

In the \textbf{singular \textit{they}} category, \textit{bad translation} and
\textit{number agreement} stand out as the most common error types, whereas the
errors in the \textbf{group \textit{it/they}} categories are distributed quite
evenly over different types. Nevertheless, the two categories exhibit similar
problems. Many of these examples involve a named entity in one sentence referred
to by the pronoun \textit{they}, usually translated as \textit{ils}, in the next
sentence.  There are two main problems, one of translation and one of
evaluation. First, it turns out that some of the named entity antecedents pose a
real challenge to the MT systems. It is very common for multi-word named
entities to be spectacularly mistranslated. For names like `Deep Mind' or
`National Ignition Facility', the MT systems produce pseudo-compositional
translations that can hardly be recognised as proper names in the output, let
alone assigned to a particular gender. As a result, the annotators struggled to
determine what pronoun should be used to refer to the entity in the next
sentence. The second problem, related to evaluation, is that the annotators
reportedly find it difficult to assess whether or not \textit{ils} can be a
correct translation of \textit{they} in these examples. Both annotators agree
that referring to the named entities in question with \textit{ils} is considered
ungrammatical in French and sounds much less natural than the literally
equivalent English pattern. Such examples are the main source of the
\textit{number agreement} errors in these categories. Still, in some cases the
annotators are reluctant to categorically label the translations as unacceptable
as this pronoun use is occasionally encountered in informal French speech. 

In the \textbf{event \textit{it}} category, \textit{wrong pronoun type} is by
far the most common source of errors. In almost all of these cases, the English
pronoun \textit{it} was translated with the French personal pronoun \textit{il}
or, occasionally, \textit{elle} or \textit{ils}. The correct choice in these
cases would usually have been the demonstrative pronoun \textit{cela} or
\textit{ça}. This confirms that pronoun choice in translation depends in a
crucial way on the function of the pronoun \cite{GuillouThesis}, and suggests
that pronoun function identification \cite{Loaiciga2017} may be useful for MT.

Finally, in the \textbf{pleonastic \textit{it}} category, \textit{wrong pronoun
type} and \textit{bad translation} are the most common error sources. However,
this error category is probably over-represented in our data set, as many of the
systems specifically manipulate pronouns in an imperfect attempt to improve
phenomena like gender agreement. In the pleonastic category, the uninformed
``default'' translation of the pronouns is often correct, and there is a great
risk of introducing errors by making changes to it. This is reflected by the
fact that the performance of the \textsc{baseline} system is among the best in
this category (see Table~\ref{table:ProResults}). The \textsc{baseline} did not
have any instances of \textit{wrong pronoun type} in the meta-evaluation, and
there were only two cases of \textit{bad translation}.






\subsection{Comparison Across MT Systems}

The data collected in our study also allows us to make a comparison across the
rule-based, SMT, and NMT systems represented in the data set. 

Our manual evaluation procedure depends on word alignments between pronouns and
their translations. For the SMT systems, word alignments were obtained directly
from the MT decoder.
For the rule-based and NMT systems, we rely on automatic word alignments
generated retrospectively using GIZA++ \cite{Och2003}. We reject the option of
using the output of the attention model for NMT systems as it is known that
attention and word alignment may dramatically diverge \cite{Knowles2017}.
Retrospectively computed alignments are clearly less accurate than those output
by a decoder. As a result, the number of examples that could not be annotated
correctly because of incorrect word alignment is very high, especially for the
\textsc{its2} and \textsc{limsi} systems
(Table~\ref{table:SupplementaryResults}).

We start by noting that only 230 out of 250 pronouns in the \textbf{reference
translation} were deemed to be correctly translated (Table \ref{table:ProResults}).  In general, the problems in the reference
translation are not very severe. They include typographical errors (e.\,g.,
\textit{elle/elles} and \textit{il/ils} are sometimes confused because they
share the same pronunciation) and a few cases where the reference structure
in the translation was subtly altered in a way that the annotators did not accept.


The bulk of the systems (\textsc{baseline}, \textsc{idiap},
\textsc{uu-Tiede\-mann}, \textsc{uu-Hardmeier}, \textsc{auto-\allowbreak
postEDIt}) are \textbf{phrase-\mbox{}based SMT} systems built using
the same technology and training data. As the shared task focused
on pronoun translation, all systems except for the \textsc{baseline} are
extended in some way to handle pronouns in specific ways. Nevertheless, none of
the systems manages to reduce the number of \textit{gender agreement} or
\textit{wrong pronoun type} problems below the \textsc{baseline}. In fact, some
of the systems achieved noticeably worse results. Compared to the rule-based and
NMT systems, a high number of examples are tagged for missing pronouns.

\textsc{its2} is the only purely \textbf{rule-based} system in the corpus. The
pronoun-related rules of the system are restricted in the sense that English
personal pronouns in subject position are always rendered with one of the French
personal pronouns \textit{il}, \textit{elle}, \textit{ils} or \textit{elles}.
The demonstratives \textit{ça}, \textit{cela} and \textit{ce} were never
produced. This strategy was not accepted by our annotators, resulting
in an extraordinarily high number of \textit{wrong pronoun type} annotations. A
similar observation can be made about \textsc{auto-postEDIt}, an SMT system
with rule-based postprocessing. Like \textsc{its2}, it relies on rules that do not produce
the full range of pronouns (avoiding \textit{ça} and \textit{cela}) and
is penalised for that in the evaluation.

The three \textbf{NMT} systems exhibit rather different performance.
\textsc{limsi} has fewer examples tagged for \textit{gender agreement} and
\textit{wrong pronoun type} errors than \textsc{nyu}. Unfortunately, this
cannot be unequivocally interpreted as evidence of better performance because
\textsc{limsi} produced a very high number of truncated sentences (labelled
\textit{missing text}), later found to be a result of poorly aligned sentence
pairs in the OpenSubtitles2016 dataset\footnote{Personal communication with
Rachel Bawden.}, and it is likely that it avoided many potential errors
simply by failing to produce any output at all.

The \textsc{yandex} system, a recent NMT system that builds on the Transformer
NMT architecture \cite{Vaswani2017} and models the current sentence together
with one previous sentence of context, performs much better on the test suite
than all the other systems in our corpus. This indicates that an up-to-date NMT
can have an edge over most previous MT technology when it comes to pronoun
translation. It is also noteworthy that the \textsc{yandex} system has much
fewer cases of \emph{wrong pronoun type} than the other systems, a fact that can
primarily be put down to much better performance on the \emph{event it}
category.

Interestingly, however, the system performs much better on the intra-sentential
anaphoric categories than the inter-sentential ones. This suggests that the
Transformer architecture has a real advantage over recurrent NMT models in
propagating agreement information within the scope of a sentence and that the
additional context (of the previous sentence) available to the \textsc{yandex}
model is not being harnessed to its full potential. In fact, 43 of the 55
instances of anaphoric inter-sentential pronouns in PROTEST have antecedents in
the previous sentence, but only 18 of these are correctly translated by the
system. This result contrasts with the reported claims of a noticeable
improvement in the translation of inter-sentential anaphoric pronouns with the
original English--Russian \textsc{yandex} system \cite{Voita2018}. A more
detailed study of this discrepancy must be left to future work.

\section{Discussion}
\label{sec:Discussion}

By re-evaluating a corpus of MT systems that had already been manually evaluated
in a different way for DiscoMT 2015, our study touches on questions of both MT
evaluation and MT performance. It also reveals some information about the
pronoun uses that are most problematic for MT. 

\subsection{MT Evaluation}

The evaluation of pronoun translation in an MT context is surprisingly
challenging. This is not immediately obvious, and in fact some earlier work on
discourse in MT focused on pronouns specifically because they were supposed to
be easier to evaluate than other aspects of discourse coherence such as lexical
choice \cite{HardmeierThesis}. Problems arise in two ways. First, pronoun usage
in corpus data is often less clear-cut than one might expect, with sometimes
vague reference and occasional violations of the rules of gender and number
agreement imposed by prescriptive grammar. Second, MT output is often disfluent
in various ways. Since pronouns have very little semantic content other than
their context-dependent referential properties, they become extremely difficult
to interpret and judge once the context is disturbed. Our comparison of the
DiscoMT gap-filling and the PROTEST test suite evaluation reveals problems of
both kinds. It also shows the danger of using non-native speakers as evaluators,
resulting in a high number of annotation errors despite best efforts.
General-purpose MT evaluation methods such as those used at WMT \cite{WMT2017}
arguably focus more on the adequacy of content words and may be more robust to
minor disfluencies. The effect of pronoun translation on general-purpose human
MT evaluation is an interesting follow-up problem for future work.

\subsection{MT Performance}


Gender agreement was long assumed to be the most important problem that needed
to be addressed to solve the issue of pronoun translation
\cite{LeNagard2010,Hardmeier2010}. This was eventually recognised to be
insufficient, and Guillou suggested that the function of pronouns was another
important factor affecting their translation \cite{GuillouThesis}. Our
evaluation results confirm that both of these factors play an important role.
In our study, gender agreement is by far the most common error type for
anaphoric pronouns. Beyond doubt, resolution and maintenance of coreference
relations are essential problems that must be tackled in MT research. At the
same time, many pronoun choice errors can be attributed to an incorrect
identification of pronoun function. These errors are especially frequent among
anaphoric pronouns with non-nominal antecedents, categorised as \textit{event}
pronouns in PROTEST. It seems, therefore, that noun-noun coreference does not
provide sufficient information for the correct translation of arbitrary
pronouns. To generate correct translations, the function of the input pronouns,
and to the extent they are identified as anaphoric, the type of anaphoric
reference they encode, must be taken into consideration. Among the systems in
our data set, the \textsc{yandex} system stands out by achieving much better
disambiguation between anaphoric and event pronouns. It is an interesting
question for future work whether this is primarily due to the Transformer
architecture or to the additional context encoder in the system, and whether
this disambiguation capability can be harnessed to improve other NLP tasks such
as coreference resolution.

So far, non-subject position pronouns have received little attention from the
SMT community. We find that the typical error patterns in this category differ
significantly from those of pronouns in subject position. The abysmal
performance of the MT systems for non-subject inter-sentential anaphora, where
only one system translated more than one item correctly, may to some extent be a
result of random variation due to the very small sample size (5 examples per
system). However, the results in the intra-sentential category are also very
low, pointing to serious difficulties in the translation of these pronouns. The
dominant cause of error for non-subject pronouns is omission in the target
language. For the phrase-based SMT systems, this may be due to unreliable
automatic word alignments of these pronouns. Word alignment is difficult both
because the French direct object pronouns are homonymous with the more frequent
definite articles and because of word reordering between post-verbal English and
pre-verbal French object pronouns. The rule-based and NMT systems do not perform
significantly better on this category, but the \textsc{yandex} system achieves
an improvement, especially in the intra-sentential case and probably due to the
Transformer architecture. Specifically for non-subject pronouns, it also
achieves better results than the other system in the inter-sentential case, but
the number of examples of this kind in PROTEST is too small to make a definitive
statement, especially since we fail to observe a corresponding improvement for
inter-sentential subject pronouns.

Our corpus contains one completely rule-based MT system (\textsc{its2}) and one SMT
system with rule-based post-editing (\textsc{auto-postEDIt}). The overall
performance of the two systems differs greatly, but both of them generate a
large number of pronouns of incorrect type (e.\,g., personal pronouns instead of
demonstratives; Table~\ref{tab:synth-systems}). This is likely because the
rule-based component of both systems emits only a restricted subset of all
possible target-language pronouns; in particular, neither system will ever
generate the French demonstrative \textit{ça/cela} for an English personal
pronoun. This suggests that the complexity of the rule-based components in the
MT systems we evaluated is insufficient.

\subsection{Pronoun Use}

Our study does not focus primarily on the linguistic constraints of pronoun use,
however, it is worth highlighting those aspects that pose particular problems for MT and its evaluation.
Whilst there is a clear correlation between the semantic plurality of an
entity and number marking on the associated linguistic expressions, this
correlation is not absolute, and quite frequently actual and
grammatical number diverge. Typical examples in English are the use of
\textit{they} referring to group nouns or to individuals of
unknown gender. The constraints governing the use of such forms are highly
language-dependent. Our annotators agree that they are distinctly less natural
in French, even though they do occur occasionally in informal speech.
For practical MT, this has two implications. First, we cannot assume that
linguistic properties such as number marking will be consistent in a coreference
chain or invariant under translation. Second, while literal translations may
sometimes work despite cross-linguistic differences in language use, they will
be perceived as unnatural especially if frequent.


\section{Conclusions}

We have evaluated the quality of pronoun translation in a corpus of translations
generated by different types of MT systems using a test suite of examples that
is balanced to cover different uses of pronouns. Our results demonstrate that
pronouns are problematic for all of the MT technologies we considered. In
particular, we find no evidence that the adoption of neural methods in MT by
itself leads to significantly better performance on this type of problem. Our
results do suggest, however, that the latest generation of Transformer-based NMT
models are better at handling cases of intra-sen\-ten\-tial anaphora and at
identifying the functional properties of pronouns. However, the advantage of a
context encoder, a major contribution of the \textsc{yandex} system, is less
clear, as the system fails to outperform previous technology even in those cases
where the required information is available in the scope of the context sentence.
It is therefore too early to suggest that NMT has solved the problem of pronoun
translation, but the results are encouraging.

We find two major sources of pronoun translation errors in our
English--French corpus. First, lacking awareness of \emph{pronoun function}
causes confusion between, primarily, personal pronouns and demonstratives in the
target language. Second, lacking awareness of the \emph{referential properties}
of the pronouns results in incorrect gender and number agreement. 
We recommend that system developers address both of these factors, as simple
heuristic approaches can demonstrably lead to decreased performance. 

We also highlight that the evaluation of pronoun translation is in itself a
difficult problem, as evidenced by the disagreement between the two manual
evaluation methods applied to the DiscoMT data set. As an alternative to fully
automatic evaluation we recommend the use of semi-automatic methods in
combination with hand-crafted test suites or challenge sets.

\bibliographystyle{IEEEtran}
\bibliography{WMT2018}

\end{document}